\documentclass[utf8]{FrontiersinHarvard} 

\usepackage{url,lineno,microtype,subcaption}
\usepackage[pagebackref=true,breaklinks=true,colorlinks,citecolor=blue,linkcolor=blue,bookmarks=false]{hyperref}
\usepackage[onehalfspacing]{setspace}
\usepackage{graphicx}
\usepackage{caption}
\usepackage{subcaption}
\usepackage{booktabs, multirow} 
\usepackage{colortbl}

\newcommand{\best}[1]{{\color{blue}{\textbf{#1}}}} 
\newcommand{\second}[1]{{\color{black}{\textbf{#1}}}}

\newcommand{\ours}{{MRET}}
\newcommand{\ourresult}[1]{{\cellcolor{Gray!20}{#1}}}

\def\keyFont{\fontsize{8}{11}\helveticabold }
\def\firstAuthorLast{Junjie Ke {et~al.}} 
\def\Authors{Junjie Ke\,$^{1,*}$, Tianhao Zhang\,$^{2}$, Yilin Wang\,$^{2}$, Peyman Milanfar\,$^{1}$, and Feng Yang\,$^{1}$}
\def\Address{$^{1}$Google Research, Mountain View, California, USA \\
$^{2}$Google, Mountain View, California, USA  }
\def\corrAuthor{Junjie Ke}

\def\corrEmail{junjiek@google.com}

\begin{document}
\onecolumn
\firstpage{1}

\title {\ours: Multi-resolution Transformer for Video Quality Assessment} 

\author[\firstAuthorLast ]{\Authors} 
\address{} 
\correspondance{} 

\extraAuth{}

\maketitle

\begin{abstract}

\noindent No-reference video quality assessment (NR-VQA) for user generated content (UGC) is crucial for understanding and improving visual experience. Unlike video recognition tasks, VQA tasks are sensitive to changes in input resolution. Since large amounts of UGC videos nowadays are 720p or above, the fixed and relatively small input used in conventional NR-VQA methods results in missing high-frequency details for many videos. In this paper, we propose a novel Transformer-based NR-VQA framework that preserves the high-resolution quality information. With the multi-resolution input representation and a novel multi-resolution patch sampling mechanism, our method enables a comprehensive view of both the global video composition and local high-resolution details. The proposed approach can effectively aggregate quality information across different granularities in spatial and temporal dimensions, making the model robust to input resolution variations. Our method achieves state-of-the-art performance on large-scale UGC VQA datasets LSVQ and LSVQ-1080p, and on KoNViD-1k and LIVE-VQC without fine-tuning.

\tiny
 \keyFont{ \section{Keywords:} video quality assessment, transformer, no-reference, multi-resolution, user-generated content} 
\end{abstract}

\begin{figure}[htp!]
\centering
\includegraphics[width=10cm]{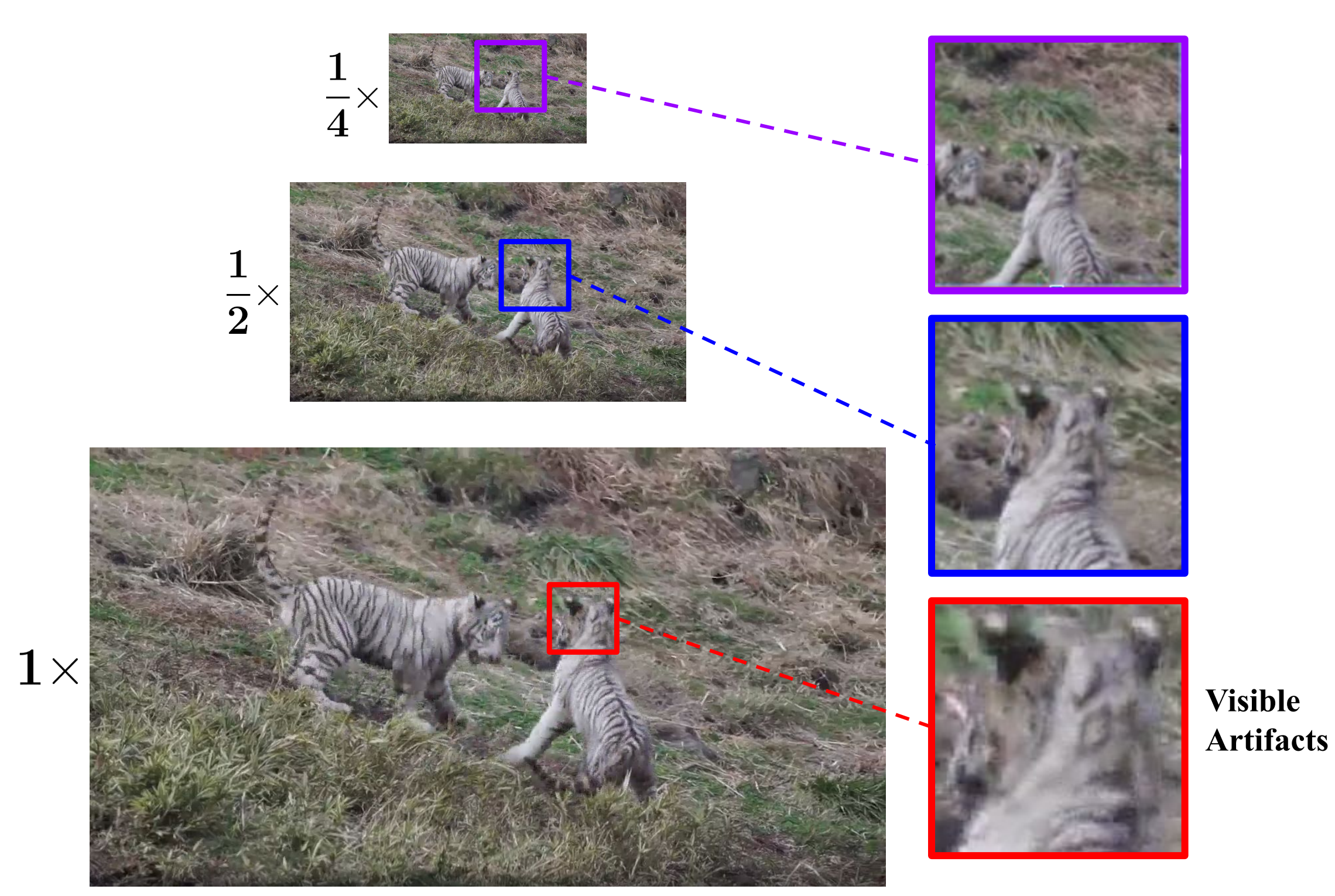}
\caption{ Video quality is affected by both global video composition and local details. Although downsampled video frames provide the global view and are easier to process for deep-learning models, some distortions visible on the original high resolution videos may disappear when resized to a lower resolution.}
\label{fig:motivation}
\end{figure}

\section{Introduction}
\begin{figure}[tp!]
\centering
\includegraphics[width=10cm]{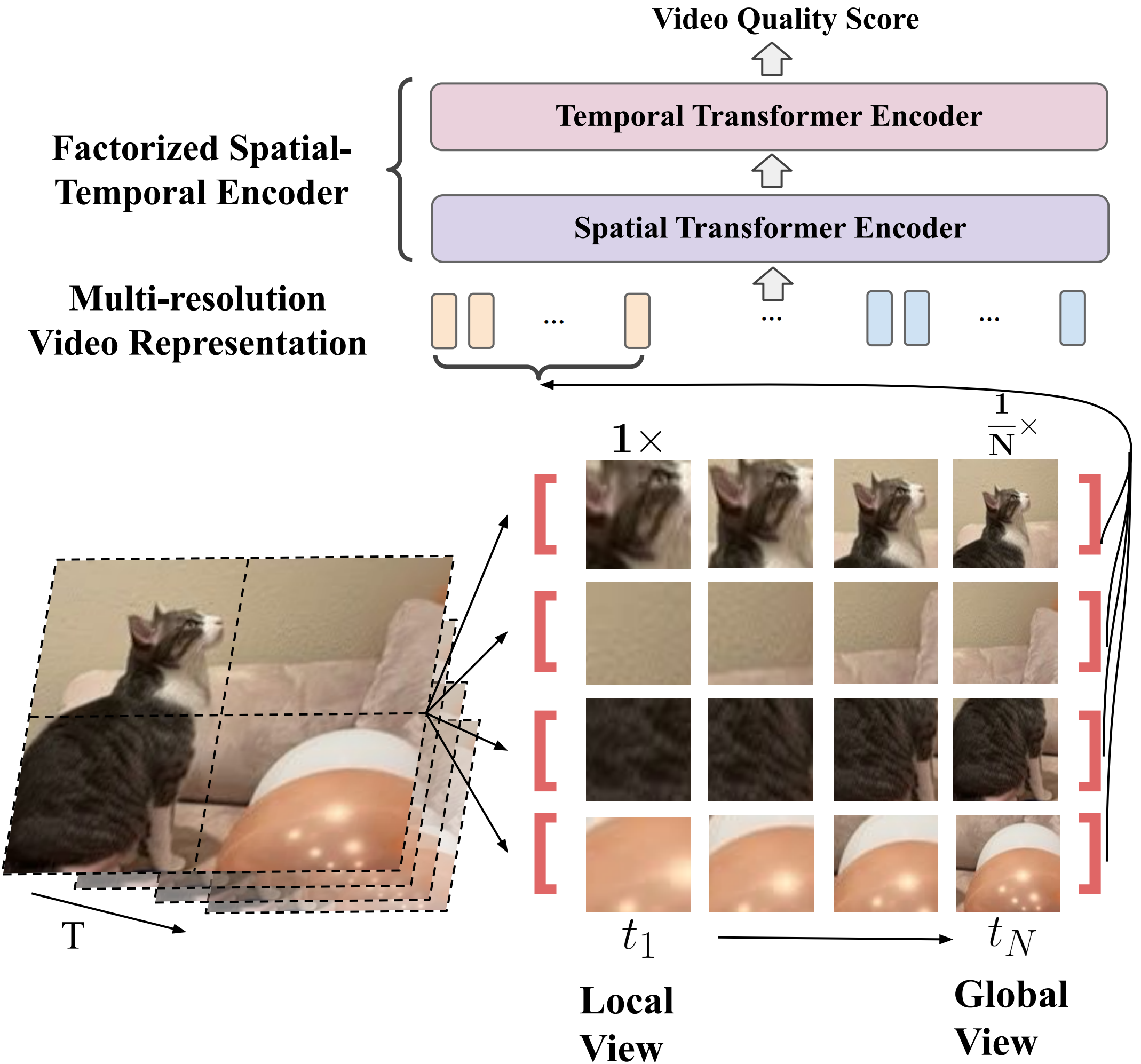}
\caption{The proposed multi-resolution Transformer (\ours) for VQA. To capture both global composition and local details of video quality, we build a multi-scale video representation with patches sampled from proportionally resized frames with different resolutions.}
\label{fig:teaser}
\end{figure}

\noindent Video quality assessment (VQA) has been an important research topic in the past years for understanding and improving perceptual quality of videos. Conventional VQA methods mainly focus on full reference (FR) scenarios where distorted videos are compared against their corresponding pristine reference. In recent years, there has been an explosion of user generated content (UGC) videos on social media platforms such as Facebook, Instagram, YouTube, and TikTok. For most UGC videos, the high-quality pristine reference is inaccessible. This results in a growing demand for no-reference (NR) VQA models, which can be used for ranking, recommending and optimizing UGC videos.

Many NR-VQA models~\citep{ying2021patch, li2019quality, tu2021ugc, you2019deep, wang2021rich} have achieved significant success by leveraging the power of deep-learning.  Most existing deep-learning approaches use convolutional neural networks (CNNs) to extract frozen frame-level features and then aggregate them in the temporal domain to predict the video quality. Since frozen frame-level features are not optimized for capturing spatial-temporal distortions, this could be insufficient to catch diverse spatial or temporal impairments in UGC videos. Moreover, predicting UGC video quality often involves long-range spatial-temporal dependencies, such as fast-moving objects or rapid zoom-in views. Since convolutional kernels in CNNs are specifically designed for capturing short-range spatial-temporal information, they cannot capture dependencies that extend beyond the receptive field~\citep{bertasius2021space}. This limits CNN models' ability to model complex spatial-temporal dependencies in UGC VQA tasks, and therefore it may not be the best choice to effectively aggregate complex quality information in diverse UGC videos.

Recently, architectures based on Transformer~\citep{vaswani2017attention} have been proven to be successful for various vision tasks~\citep{arnab2021vivit, dosovitskiy2021an, carion2020end, chen2021pre}, including image quality assessment \citep{ke2021musiq}.  Unlike CNN models that are constrained by limited receptive fields, Transformers utilize the multi-head self-attention operation which attends over all elements in the input sequence. As a result, Transformers can capture both local and global long-range dependencies by directly comparing video quality features at all space-time locations. This inspires us to apply Transformer on VQA in order to effectively model the complex space-time distortions in UGC videos.

Despite the benefits of Transformers, directly applying Transformers on VQA is challenging because VQA tasks are resolution-sensitive. Video recognition models like ViViT~\citep{arnab2021vivit} use fixed and relatively small input size,~\emph{e.g.}, 224$\times$224. This is problematic for VQA since UGC videos with resolution smaller than 224 are very rare nowadays (less than 1\% in LSVQ~\citep{ying2021patch}). Such downsampling leads to missing high-frequency details for many videos. As shown in Figure~\ref{fig:motivation}, some visible artifacts in the high resolution video are not obvious when the video is downsampled. Human perceived video quality is affected by both the global video composition,~\emph{e.g.}, content, video structure and smoothness and local details,~\emph{e.g.}, texture and distortion artifacts. 
But it is hard to capture both global and local quality information when using fixed resolution inputs. Similarly for image quality assessment, \cite{ke2021musiq} showed the benefit of applying the Transformer architecture on the image at the original resolution. Although processing the original high-resolution input is affordable for a single image, it is computationally infeasible for videos, due to Transformer's quadratic memory and time complexity. 

To enable high-resolution views in video Transformers for a more effective VQA model, we propose to leverage the complementary nature of low and high resolution frames. We use the low-resolution frames for a complete global composition view, and sample spatially aligned patches from the  high-resolution frames to complement the high-frequency local details. The proposed Multi-REsolution Transformer (\ours) can therefore efficiently extract and encode the multi-scale quality information from the input video. This enables more effective aggregation of both global composition and local details of the video to better predict the perceptual video quality.

As illustrated in Figure~\ref{fig:teaser}, we first group the neighboring frames to build a multi-resolution representation composed of lower-resolution frames and higher-resolution frames. We then introduce a novel and effective multi-resolution patch sampling mechanism to sample spatially aligned patches from the multi-resolution frame input. These multi-resolution patches capture both the global view and local details at the same location, and they serve as the multi-resolution input for the video Transformer. In addition to preserving high-resolution details, our proposed \ours\ model also aligns the input videos at different resolutions, making the model more robust to resolution variations. After the multi-resolution tokens are extracted, a factorized spatial and temporal encoder is employed to efficiently process the large number of spatial-temporal tokens. 

The major contributions of this paper are summarized into three folds:

\begin{itemize}
\vspace{-2mm}
    \item We propose a multi-resolution Transformer for video quality assessment (\ours), which makes it possible to preserve high-resolution quality information for UGC VQA.
    
    \item We propose a novel multi-resolution patch sampling mechanism, enabling the Transformer to efficiently process both global composition information and local high-resolution details.

    \item We apply \ours\ on large-scale UGC VQA datasets. It outperforms the previous state-of-the-art methods on LSVQ~\citep{ying2021patch} and LSVQ-1080p~\citep{ying2021patch}. It also achieves state-of-the-art performance on KoNViD-1k~\citep{hosu2017konstanz} and LIVE-VQC~\citep{sinno2018large} without fine-tuning, demonstrating its robustness and generalization capability.
\vspace{-1mm}
\end{itemize}

\section{Related Work}

\noindent\textbf{Video Quality Assessment:} Video quality assessment aims to quantify video quality. FR-VQA methods measure quality changes from pristine videos, and NR-VQA methods measure video quality without a pristine reference. For UGC videos that lack high-quality pristine reference, NR-VQA metrics are more applicable. Conventional NR metrics \citep{saad2014blind, mittal2015completely, li2016spatiotemporal, sinno2019spatio, dendi2020no, korhonen2019two, tu2021ugc} utilize distortion-specific features and low-level features like natural scene statistics (NSS). These feature-based NR-VQA methods  mainly rely on hand-crafted statistical features summarized from limited data and are harder to generalize to diversified UGC videos. In the past few years, CNN-based NR metrics \citep{ying2021patch, li2019quality, you2019deep, wang2021rich} achieve great success in VQA using features extracted with CNNs. The features are then aggregated temporally with pooling layers or recurrent units like LSTM.  The PVQ~\citep{ying2021patch} method learns to model the relationship between local video patches and the global original UGC video. It shows that exploiting both global and local information can be beneficial for VQA.  Recent CNN-Transformer hybrid methods \citep{you2021long, tan2021no, li2021full, jiang2021multi} show the benefit of using Transformer for temporal aggregation on CNN-based frame-level features. Since all these methods use CNN for spatial feature extraction, they suffer from CNN's limitation,~\emph{i.e.}, a relatively small spatial receptive field. Moreover, these frame-level features are usually extracted from either fixed size inputs or a frozen backbone without VQA optimization. Our method is a pure Transformer-based VQA model and can be optimized end-to-end. Unlike models that use fixed small input, our proposed \ours\ model enables high-resolution inputs. The proposed multi-resolution input representation allows the model to have a full spatial receptive field across multiple scales.

\noindent\textbf{Vision Transformers:} The Transformer~\citep{vaswani2017attention} architecture was first proposed for NLP tasks and has recently been adopted for various computer vision tasks~\citep{arnab2021vivit, dosovitskiy2021an, carion2020end, chen2021pre, ke2021musiq}. The Vision Transformer (ViT)~\citep{dosovitskiy2021an} first proposes to classify an image by treating it as a sequence of patches. This seminal work has inspired subsequent research to adopt Transformer-based architectures for other vision tasks. For video recognition, ViViT~\citep{arnab2021vivit} examines four designs of spatial and temporal attention for the pretrained ViT model. TimeSformer~\citep{bertasius2021space} studies five different space-time attention methods and shows that a factorized space-time attention provides better speed-accuracy tradeoff. Video Swin Transformer~\citep{liu2022video} extends the local attention computation of Swin Transformer~\citep{liu2021swin} to temporal dimension, and it achieves state-of-the-art accuracy on a broad range of video recognition benchmarks such as Kinetics-400~\citep{kay2017kinetics} and Kinetics-600~\citep{kay2017kinetics}. Since video recognition tasks are less sensitive to input resolution than VQA, most of the video Transformers proposed for video recognition tasks use relatively small resolution and fixed square input,~\emph{e.g.} $224 \times 224$. The objective for the VQA task is sensitive to both global composition and local details, and it motivates us to enable video Transformers to process frames in a multi-resolution manner, capturing both global and local quality information. 

\section{Multi-resolution Transformer for Video Quality Assessment}

\subsection{Overall Architecture}
\label{sec:overall}

\noindent Understanding the quality of UGC videos is hard because they are captured under very different conditions like unstable cameras, imperfect camera lens, varying resolutions and frame rates, different algorithms and parameters for processing and compression. As a result, UGC videos usually contain a mixture of spatial and temporal distortions. Moreover, the way viewers perceive the content and distortions also impact the perceptual quality of the video. Sometimes transient distortions such as sudden glitches and defocusing can significantly impact the overall perceived quality, which makes the problem even more complicated. As a result, both global video composition and local details are important for accessing the quality of UGC videos.  

\begin{figure*}[tp!]
\centering
\centering
\includegraphics[width=17cm]{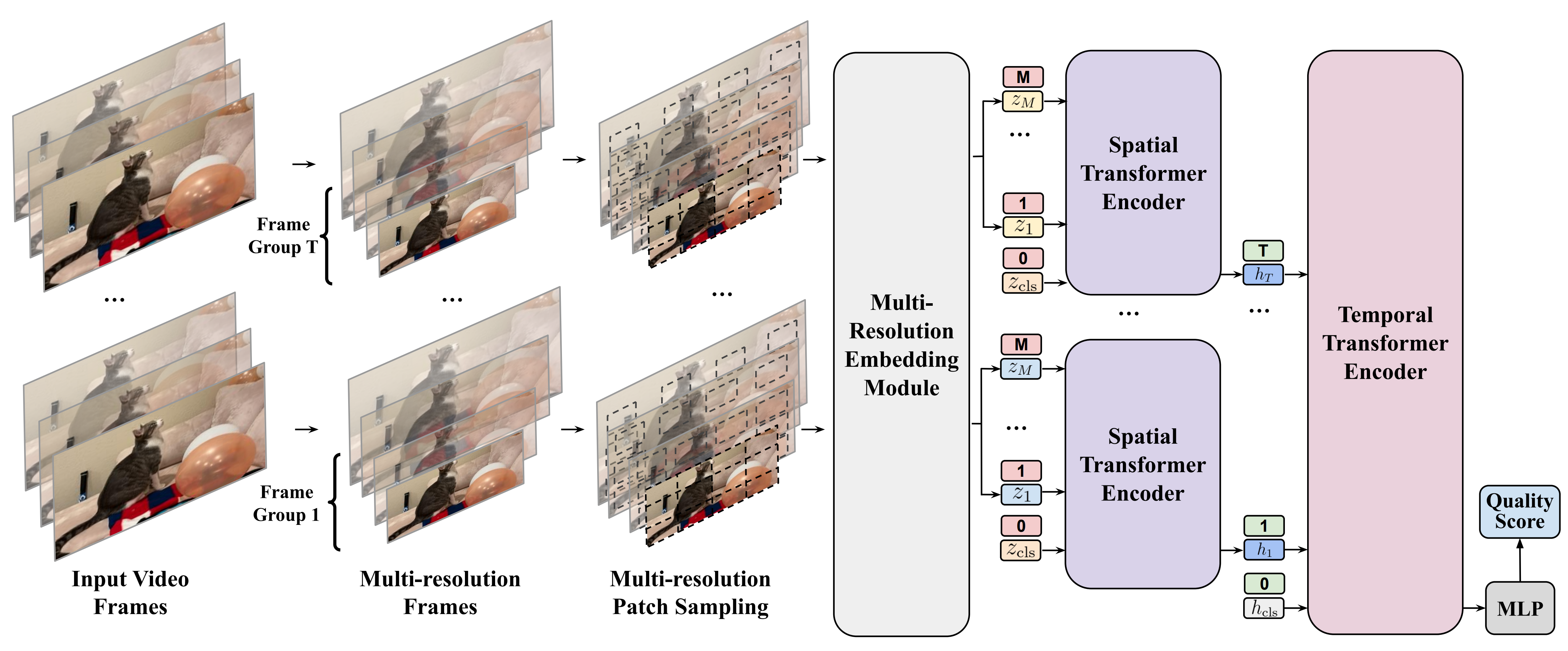}
\caption{Model overview for \ours. Neighboring video frames are grouped and rescaled into a pyramid of low-resolution and high-resolution frames. Patches are sampled from the multi-resolution frames and encoded as the Transformer input tokens. The spatial Transformer encoder takes the multi-resolution tokens to produce a representation per frame group at its time step. The temporal Transformer encoder then aggregates across time steps. To predict the video quality score, we follow a common strategy in Transformers to prepend a ``classification token" ($z_{cls}$ and $h_{cls}$) to the sequence to represent the whole sequence input and to use its output as the final representation.}
\label{fig:overview}
\end{figure*}

To capture video quality at different granularities, we propose a multi-resolution Transformer (\ours) for VQA which embeds video clips as multi-resolution patch tokens as shown in  Figure~\ref{fig:overview}. \ours\ is comprised of two major parts, namely 1) a multi-resolution video embedding module (Section \ref{sec:multi-resolution-representation}), and 2) a space-time factorized Transformer encoding module (Section \ref{sec:factorized-encoder}). 

The multi-resolution video embedding module aims to encode the multi-scale quality information in the video, capturing both global video composition from lower resolution frames, and local details from higher resolution frames. The space-time factorized Transformer encoding module aggregates the spatial and temporal quality from the multi-scale embedding input.

\subsection{Multi-resolution Video Representation}
\label{sec:multi-resolution-representation}

\begin{figure}[!tp]
\centering
\captionsetup{font=small}
\includegraphics[width=10cm]{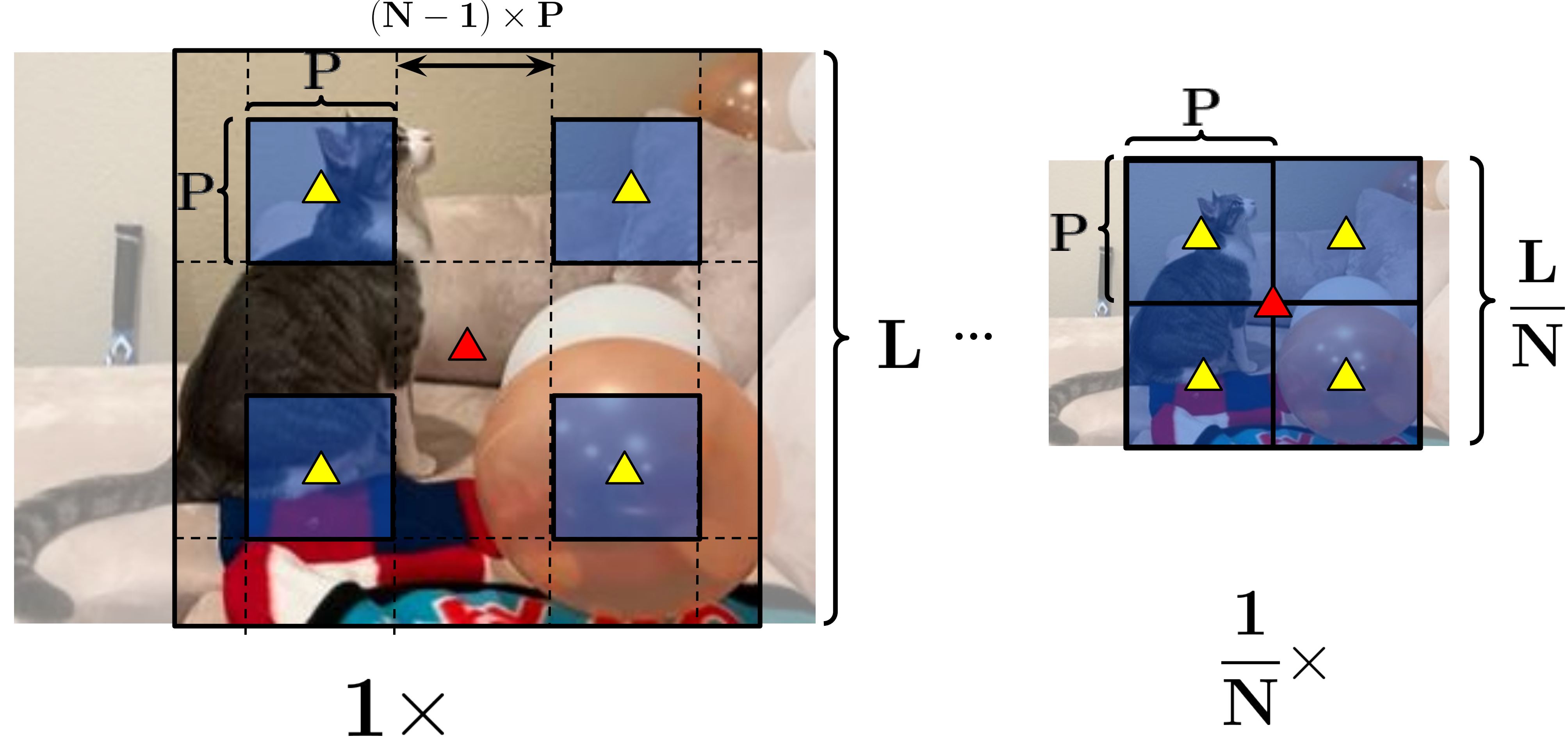}
\caption{Multi-resolution patch sampling. We first rescale the $N$ frames to $L, ..., \frac{2L}{N}, \frac{L}{N}$ for the shorter side and uniformly sample grid of patches from the multi-resolution frames. $P$ is the patch size. Patches are spatially aligned. The patches at the same location in the grid provide a multi-scale view for the same location.}
\label{fig:patch-sampling}
\end{figure}

\begin{figure}[!tp]
\centering
\captionsetup{font=small}
\includegraphics[width=10cm]{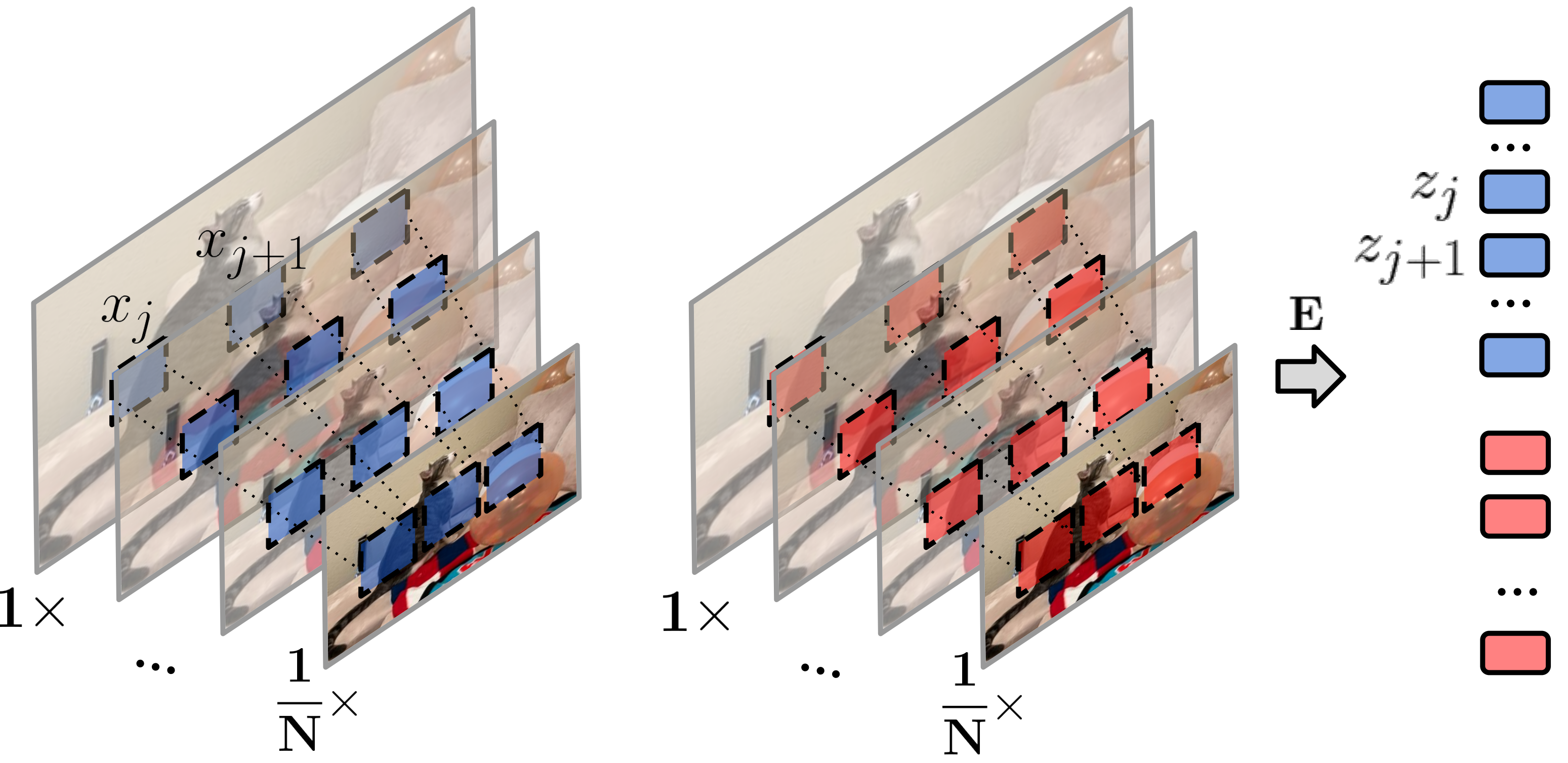}
\caption{Multi-resolution video frames embedding. We extract center-aligned multi-resolution patches, and then linearly project the spatially aligned ``tubes" of patches to 1D tokens.}
\label{fig:tubelet}
\end{figure}

\noindent Since UGC videos are highly diverse, we need to design an effective multi-resolution video representation for capturing the complex global and local quality information. To achieve that, we first transform the input video into groups of multi-resolution frames. As shown in Figure~\ref{fig:overview},  the input frames are divided into groups of $N$. $N$ is the number of scales in the multi-resolution input. We then resize the $N$ frames into a pyramid of low-resolution and high-resolution frames. We preserve the aspect ratios of the frames during resizing, and we control the shorter-side length for each frame (Figure~\ref{fig:patch-sampling}). Assuming the shorter-side length for the largest resolution is $L$, the resulting pyramid of frames will have shorter-side length $L, ..., \frac{2L}{N}, \frac{L}{N}$ accordingly. As a result, we will have a pyramid of $N$ frames, scaling from $1\times$ to  $\frac{1}{N}\times$ resolution.

After obtaining the multi-resolution frames, we need a way to effectively and efficiently encode them as input tokens to the Transformer. Although low-resolution frames can be processed efficiently, processing the high-resolution frames in its entirety can be computationally expensive. For the higher-resolution frames, we propose to sample patches instead to save computation. Intuitively, the lower-resolution frames provide global views of the video composition, while the higher-resolution ones provide complementary local details. We want a patch sampling method that can best utilize the complementary nature of these multi-scale views. To achieve that, we propose to sample spatially aligned grids of patches from the grouped multi-resolution frames. In short, we use the lowest resolution frame for a complete global view, and we sample local patches at the same location from the higher-resolution frames to provide the multi-scale local details. Since the patches are spatially aligned, the Transformer has access to both the global view and local details at the same location. This allows it to better utilize the complementary multi-scale information for learning video quality.

Figure~\ref{fig:patch-sampling} and Figure~\ref{fig:tubelet} demonstrate how we sample spatially aligned grids of patches. Firstly, we choose a frame center, as shown by the red triangle in Figure~\ref{fig:patch-sampling}. During training, the frame center is chosen randomly along the middle line for the longer-length side. For inference, we use the center of the video input. After aligning the frames, we then sample center-aligned patches from the frames. $P$ is the patch size. For the smallest frame, we continuously sample the grid of patches to capture the complete global view. For larger frames, we sample linearly spaced-out patches to provide multi-scale local details. The center for the patches remain aligned at the same location, as shown by the yellow triangles in Figure~\ref{fig:patch-sampling}. For the $i$-th frame ($i = 1, ..., N$), the distance between patches can be calculated as $(N - i)\times P$.  Since the patches are center-aligned, they form a ``tube" of multi-resolution patches for the same location.  As a result, those multi-resolution patches provide a gradual ``zoom-out" view, capturing both the local details and global view at the same location. 

As shown in Figure~\ref{fig:tubelet}, we then linearly project each tube of multi-resolution patch $x_i$ to a 1D token $z_i \in \mathbb{R}^d$ using learned matrix $\mathbf{E}$ where $d$ is the dimension of the Transformer input tokens. This can be implemented using a 3D convolution with kernel size $N\times P\times P$. Each embedded token contains multi-resolution patches at the same location, allowing the model to utilize both global and local spatial quality information. Moreover, the multi-scale patches also fuse local spatio-temporal information together during tokenization. Therefore, it provides a comprehensive representation for the input video.

\subsection{Factorized Spatial Temporal Transformer}
\label{sec:factorized-encoder}

\noindent As shown in Figure~\ref{fig:overview}, after extracting the multi-resolution frame embedding, we apply a factorization of spatial and temporal Transformer encoders in series to efficiently encode the space-time quality information. Firstly, the spatial Transformer encoder takes the tokens from each frame group to produce a latent representation per frame group. It serves as the representation at this time step. Secondly, the temporal Transformer encoder models temporal interaction by aggregating the information across time steps.

\subsubsection{Spatial Transformer Encoder}
\noindent The spatial Transformer encoder aggregates the multi-resolution patches extracted from the entire frame group to a representation $h_t \in \mathbb{R}^d$ at its time step where $t = 1, ..., T$ is the temporal index for the frame group. $T$ is the number of frame groups. As mentioned in the previous section, for multi-resolution patches $x_i$ from each frame group, we project it to a sequence of multi-resolution tokens as $z_i \in \mathbb{R}^d$, $i = 1, ..., M$ using learnable matrix $\mathbf{E}$ where $M$ is the total number of patches. We follow the standard approach of prepending an extra learnable ``classification token" ($z_{cls} \in \mathbb{R}^d$) \citep{devlin2018bert, dosovitskiy2021an} and use its representation at the final encoder layer as the final spatial representation for the frame group. Additionally, a learnable 
spatial positional embedding $\mathbf{p} \in \mathbb{R}^{M\times d}$ is added element-wisely to the input tokens $z_i$ to encode spatial position. The tokens are passed through a Transformer encoder with $K$ layers. Each layer $k$ consists of multi-head self-attention (MSA), layer normalization (LN), and multilayer perceptron (MLP) blocks. The spatial Transformer encoder is formulated as:

\vspace{-4mm}
\begin{align}
    &\mathbf{z}_0 = [z_{cls}, \mathbf{E} x_1, \mathbf{E} x_2, ..., \mathbf{E} x_M] + \mathbf{p} \\
    &\mathbf{z}^{\prime}_k = \text{MSA}(\text{LN}(\mathbf{z}_{k-1})) + \mathbf{z}_{k-1},\ \ \ \ \ \ \ \ \ \  k =1\cdots K \\
    &\mathbf{z}_k = \text{MLP}(\text{LN}(\mathbf{z}^{\prime}_{k})) + \mathbf{z}^{\prime}_{k},\ \ \ \ \ \ \ \ \ \ \ \ \ \ \ \ \ \ k =1\cdots K \\
    &h_t = \text{LN}(\mathbf{z}_K^0) &
\vspace{-4mm}
\end{align}

\subsubsection{Temporal Transformer Encoder}
\noindent The temporal Transformer encoder models the interactions between tokens from different time steps. We use the $z_{cls}$ token position output from the spatial Transformer encoder as the frame group level representation. As shown in Figure \ref{fig:overview}, each group of frames will be encoded as a single token $h_t$, $t = 1, ..., T$. We then prepend a $h_{cls} \in \mathbb{R}^d$ token and add a separate learnable temporal positional embedding $\mathbf{p^t} \in \mathbb{R}^{T\times d}$ to the temporal tokens. These tokens are then fed to the temporal Transformer encoder, which models the temporal interactions across time. The output at the $h_{cls}$ token is used as the final representation for the whole video. Similarly, the temporal Transformer encoder can be formulated as:
\begin{align}
    &\mathbf{h}_0 = [h_{cls}, h_1, h_2, ..., h_T] + \mathbf{p^t} \\
    &\mathbf{h}^{\prime}_q = \text{MSA}(\text{LN}(\mathbf{h}_{q-1})) + \mathbf{h}_{q-1},\ \ \ \ \ \ \ \ \ \  q =1\cdots Q \\
    &\mathbf{h}_q = \text{MLP}(\text{LN}(\mathbf{h}^{\prime}_{q})) + \mathbf{h}^{\prime}_{q},\ \ \ \ \ \ \ \ \ \ \ \ \ \ \ \ \ \ q =1\cdots Q \\
    &v = \text{LN}(\mathbf{h}_Q^0) &
\end{align}
$Q$ is the number of layers for the temporal Transformer encoder. $v$ is output from the $h_{cls}$ token position from the temporal encoder, which is used as the final video representation.

\subsection{Video Quality Prediction}
\label{sec:vqa-prediction}

\noindent To predict the final quality score, we add an MLP layer on top of the final video representation $v$. The output of the MLP layer is regressed to the video mean opinion score (MOS) label associated with each video in VQA datasets. The model is trained end-to-end with $L_2$ loss.

\subsection{Initialization from Pretrained Models}
\label{sec:initialization}

\noindent Vision Transformers have been shown to be only effective when trained on large-scale datasets~\citep{dosovitskiy2021an, arnab2021vivit} as they lack the inductive biases of 2D image structures, which needs to be imposed during pretraining. However, existing video quality datasets are several magnitudes smaller than large-scale image classification datasets, such as ILSVRC-2012 ImageNet~\citep{imagenet} (we refer to it as ImageNet in what follows) and  ImageNet-21k~\citep{deng2009imagenet}. As a result, training Transformer models from scratch using VQA datasets is extremely challenging and impractical. We therefore also choose to initialize the Transformer backbone from pretrained image models.

Unlike the 3D video input, the image Transformer models only need 2D projection for the input data. To initialize the 3D convolutional filter $\mathbf{E}$ from 2D filters $\mathbf{E}_{\text{image}}$ in pretrained image models, we adopt the ``central frame initialization strategy" used in ViViT~\citep{arnab2021vivit}. In short, $\mathbf{E}$ is initialized with zeros along all temporal positions, except at the center $\lfloor{N / 2}\rfloor$. The initialization of $\mathbf{E}$ from pretrained image model can therefore be formulated as:

\vspace{-2mm}
\begin{equation}
\mathbf{E} = [\mathbf{0}, ..., \mathbf{E}_{\text{image}}, ..., \mathbf{0}]
\vspace{-2mm}
\end{equation}

\section{Experimental Results}

\subsection{Datasets}

\noindent We run experiments on four UGC VQA datasets, including LSVQ~\citep{ying2021patch}, LSVQ-1080p~\citep{ying2021patch}, KoNViD-1k~\citep{hosu2017konstanz}, and LIVE-VQC~\citep{sinno2018large}. LSVQ (excluding LSVQ-1080p) consists of 38,811 UGC videos and 116,433 space-time localized video patches. The original and patch videos are all annotated with MOS scores in [0.0, 100.0], and it contains videos of diverse resolutions. LSVQ-1080p contains 3,573 videos with  1080p resolution or higher. Since our model does not make a distinction between original videos and video patches, we use all the 28.1k videos and 84.3k video patches from the LSVQ training split to train the model and evaluate the model on full-size videos from the testing splits of LSVQ and LSVQ-1080p. KoNViD-1k contains 1,200 videos with MOS scores in [0.0, 5.0] and 960p fixed resolution. LIVE-VQC contains 585 videos with MOS scores in [0.0, 100.0] and video resolution from 240p to 1080p. We use KoNViD-1k and LIVE-VQC for evaluating the generalization ability of our model without fine-tuning. Since no training is involved, we use the entire dataset for evaluation.

\subsection{Implementation Details}
\noindent We set the number of multi-resolution frames in each group to $N=4$. The shorter-side length $L$ is set to 896 for the largest frame in the frame group. Correspondingly, the group of frames are rescaled with shorter-side length 896, 672, 448, and 224. We use patch size $P=16$ when generating the multi-resolution frame patches. For each frame, we sample a $14\times 14$ grid of patches. Unless otherwise specified, the input to our network is a video clip of 128 frames uniformly sampled from the video. 

The hidden dimension for Transformer input tokens is set to $d=768$. For the spatial Transformer, we use the ViT-Base~\citep{dosovitskiy2021an} model (12 Transformer layers with 12 heads and 3072 MLP size), and we initialize it from the checkpoint trained on ImageNet-21K~\citep{deng2009imagenet}. For the temporal Transformer, we use 8 layers with 12 heads, and 3072 MLP size. The final model has 144M parameters and 577 GFLOPs.

We train the models with the synchronous SGD momentum optimizer, a cosine decay learning rate schedule from 0.3 and a batch size of 256 for 10 epochs in total. All the models are trained on TPUv3 hardware. Spearman rank ordered correlation (SRCC) and Pearson linear correlation (PLCC) are reported as performance metrics.

\subsection{Comparison with the State-of-the-art}

\begin{table}[!tp]
\centering
\small
\begin{tabular}{lcc|ccc}\toprule
&\multicolumn{2}{c|}{LSVQ} &\multicolumn{2}{c}{LSVQ-1080p} \\\cmidrule{2-5}
Models &SRCC &PLCC &SRCC &PLCC \\\midrule
BRISQUE~\citep{mittal2012no} &0.576 &0.576 &0.497 &0.531 \\
TLVQM~\citep{korhonen2019two} &0.772 &0.774 &0.589 &0.616 \\
VIDEVAL~\citep{tu2021ugc} &0.794 &0.783 &0.545 &0.554 \\
VSFA~\citep{li2019quality} &0.801 &0.796 &0.675 &0.704 \\
PVQ~\citep{ying2021patch} &\second{0.827} &\second{0.828} &\second{0.711} &\second{0.739} \\
\midrule
\ours{} (Ours) &\ourresult{\best{0.867}} &\ourresult{\best{0.865}} &\ourresult{\best{0.780}} &\ourresult{\best{0.817}} \\
\bottomrule
\end{tabular}
\caption{Results on full-size videos in LSVQ and LSVQ-1080p test sets. Blue and black numbers in bold represent the best and second best respectively. We take numbers from \citep{ying2021patch} for the results of the reference methods. Our final method is marked in \noindent\colorbox{gray!30}{\makebox[1em]{gray}}.} \label{tab:lsvq-results}
\end{table}

\begin{table}[!tp]
\centering
\small
\begin{tabular}{clccccc}\toprule
& &\multicolumn{2}{c}{LIVE-VQC} &\multicolumn{2}{c}{KoNViD-1k} \\\cmidrule{3-6}
&Models &SRCC &PLCC &SRCC &PLCC \\\midrule
\multirow{3}{*}{w/ Fine-tune} &\cite{tan2021no} &0.760 &0.795 &0.798 &0.797 \\
&\cite{jiang2021multi} &\best{0.776} &0.789 &0.789 &0.788 \\
&LSCT-PHIQNet~\citep{you2021long} &- &- &\best{0.85} &\best{0.86} \\
&TPQI~\citep{liao2022exploring} &0.718 &0.730 &0.693 &0.693 \\\midrule
\multirow{8}{*}{w/o Fine-tune} &BRISQUE~\citep{mittal2012no} &0.524 &0.536 &0.646 &0.647 \\
&TLVQM~\citep{korhonen2019two} &0.670 &0.691 &0.732 &0.724 \\
&VIDEVAL~\citep{tu2021ugc} &0.630 &0.640 &0.751 &0.741 \\
&VSFA~\citep{li2019quality} &0.734 &0.772 &0.784 &0.794 \\
&PVQ~\citep{ying2021patch} &\second{0.770} &\second{0.807} &0.791 &0.795 \\
&LSCT-PHIQNet~\citep{you2021long} &0.737 &0.762 &- &- \\
&\ours{} (Ours) &\ourresult{\best{0.776}} &\ourresult{\best{0.817}} &\ourresult{\second{0.846}} &\ourresult{\second{0.854}} \\
\bottomrule
\end{tabular}
\caption{Performance on KoNViD-1k and LIVE-VQC. Methods except LSCT-PHIQNet~\citep{you2021long} in ``w/o Fine-tune" group are trained on LSVQ. Blue and black numbers in bold represent the best and second best respectively. We take numbers from \citep{ying2021patch, jiang2021multi, you2021long, tan2021no, liao2022exploring} for the results of the reference methods. Our final method is marked in \noindent\colorbox{gray!30}{\makebox[1em]{gray}}.} \label{tab:cross-dataset-results}
\end{table}

\begin{table*}[!tp]
\centering
\small
\begin{tabular}{cccc|cc|cc|ccc}\toprule
& &\multicolumn{4}{c|}{LSVQ} &\multicolumn{4}{c}{LSVQ-1080p} \\\cmidrule{3-10}
& &\multicolumn{2}{c|}{\ours} &\multicolumn{2}{c|}{ w/o Multi-resolution} &\multicolumn{2}{c|}{\ours} &\multicolumn{2}{c}{ w/o Multi-resolution} \\\cmidrule{3-10}
\# Frames &GFLOPs &SRCC &PLCC &SRCC &PLCC &SRCC &PLCC &SRCC &PLCC \\\midrule
32 &144 &0.844 &0.841 &0.828 &0.828 &0.749 &0.788 &0.726 &0.759 \\
64 &289 &0.857 &0.854 &0.845 &0.845 &0.768 &0.807 &0.737 &0.784 \\
96 &433 &\second{0.862} &\second{0.860} &0.851 &0.851 &\second{0.776} &\second{0.813} &0.754 &0.771 \\
128 &577 &\ourresult{\best{0.867}} &\ourresult{\best{0.865}} &0.852 &0.851 &\ourresult{\best{0.780}} &\ourresult{\best{0.817}} &0.749 &0.782 \\
\bottomrule
\end{tabular}
\caption{Ablation study results for multi-resolution input on LSVQ and LSVQ-1080p dataset. \ours\ uses multi-resolution input while ``w/o Multi-resolution" uses fixed-resolution frames. Both models grouped the frames by $N=4$ when encoding video frames into tokens. Blue and black numbers in bold represent the best and second best respectively on the same dataset. Our final method is marked in \noindent\colorbox{gray!30}{\makebox[1em]{gray}}.} \label{tab:multi-resolution-results}
\end{table*}

\subsubsection{Results on LSVQ and LSVQ-1080p} 
\noindent Table~\ref{tab:lsvq-results} shows the results on full-size LSVQ and LSVQ-1080p datasets. Our proposed \ours\ outperforms other methods by large margins on both datasets. Notably, on the higher resolution test dataset LSVQ-1080p, our model is able to outperform the strongest baseline by \textbf{7.8\%} for PLCC (from 0.739 to 0.817). This shows that for high-resolution videos, the proposed multi-resolution Transformer is able to better aggregate local and global quality information for a more accurate video quality prediction.

\subsubsection{Performance on Cross Dataset}
\noindent Since existing VQA datasets are magnitudes smaller than popular image classification datasets, VQA models are prone to overfitting. Therefore, it is of great interest to obtain a VQA model that can generalize across datasets. To verify the generalization capability of \ours, we conduct a cross-dataset evaluation where we train the model using LSVQ training set and separately eval on LIVE-VQC and KoNViD-1k without fine-tuning. As shown in Table~\ref{tab:cross-dataset-results}, \ours\ is able to generalize very well to both datasets, and it performs the best among methods \textbf{without fine-tuning}. Moreover, its performance is even as good as the best ones that are fine-tuned on the target dataset. This demonstrates the strong generalization capability of \ours. Intuitively, the proposed multi-resolution input aligns the videos at different resolutions. Not only does it provide a more comprehensive view of the video quality, but it also makes the model more robust to resolution variations.  As a result, \ours\ can learn to capture quality information for UGC videos under different conditions.

\subsection{Ablation Studies}
\subsubsection{Spatial Temporal Quality Attention}
\begin{figure}[!tp]
\centering
\captionsetup{font=small}
\includegraphics[width=15cm]{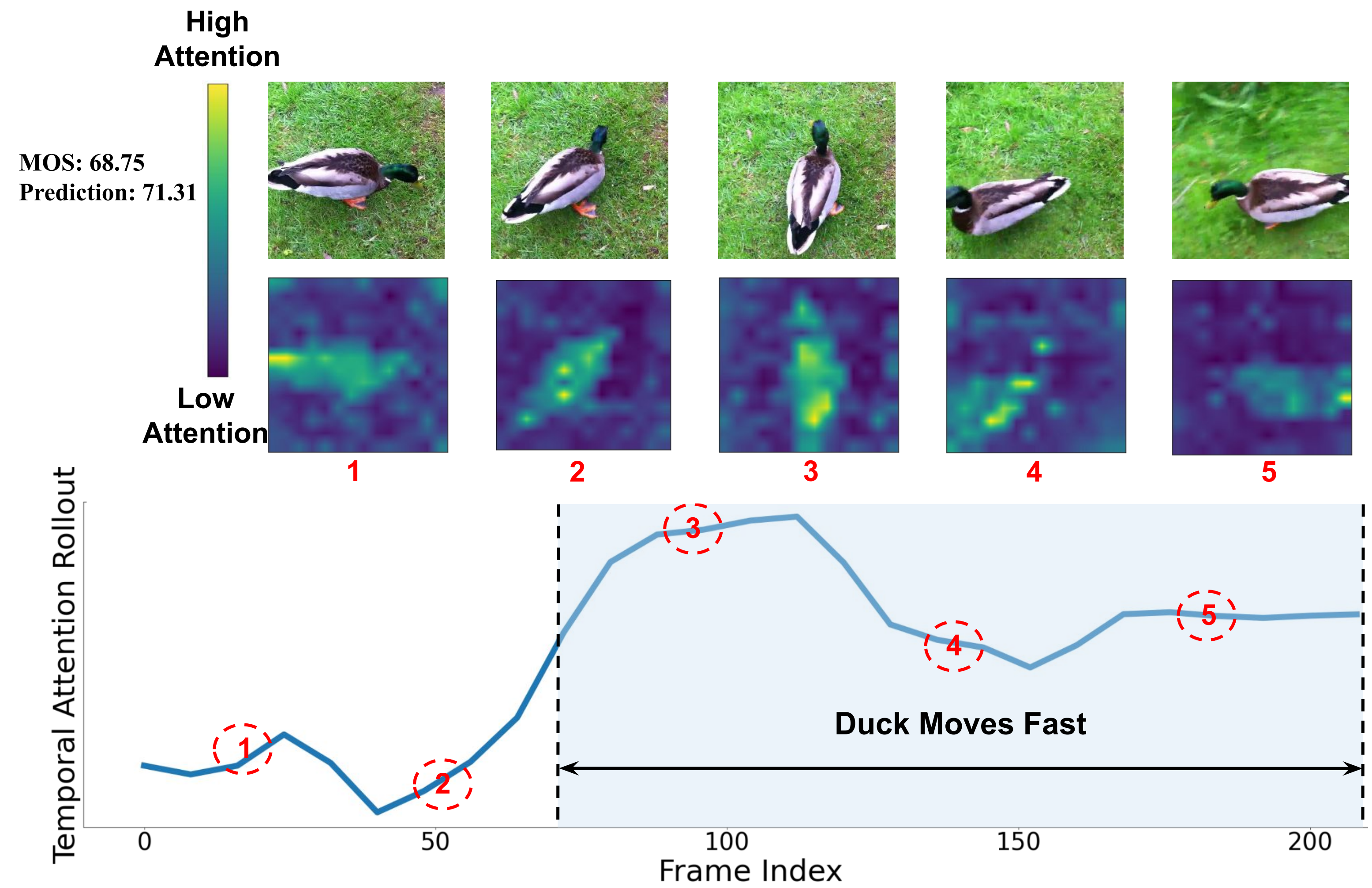}
\caption{ Visualization of spatial and temporal attention from output tokens to the input. The heat-map on the top shows the spatial attention. The chart on the bottom shows the temporal attention. Higher attention values correspond to the more important video segments and spatial regions for prediction. The model is focusing on spatially and temporally more meaningful content when predicting the final video quality score.} \label{fig:attention_visualization}
\end{figure}

\noindent To understand how \ours\ aggregates spatio-temporal information to predict the final video quality, we visualize the attention weights on spatial and temporal tokens using Attention Rollout~\citep{abnar2020quantifying}. In short, we average the attention weights of the Transformer across all heads and then recursively multiply the weight matrices of all layers. Figure~\ref{fig:attention_visualization} visualizes temporal attention for each input time step and spatial attention for selected frames. As shown by temporal attention for the video, the model is paying more attention to the second section when the duck is moving rapidly across the grass. The spatial attention also shows that the model is focusing on the main subject,~\emph{i.e.}, duck in this case. This verifies that \ours\ is able to capture spatio-temporal quality information and utilize it to predict the video quality.

\subsubsection{Effectiveness of Multi-resolution Frame Inputs} 

\noindent To verify the effectiveness of the proposed multi-resolution input representation, we run ablations by not using the multi-resolution input. The comparison result is shown in Table~\ref{tab:multi-resolution-results} as ``\ours" and ``w/o Multi-resolution" for with and without the multi-resolution frames respectively. For \ours, we resize the frames to [896, 672, 448, 224] for shorter-side lengths. For the method ``w/o Multi-resolution", we resize all the frames in the frame group to the same shorter-side length (224). The GFLOPs is the same for both models because the patch size and number of patches are the same. The multi-resolution frame input brings \textbf{1-2\%} boost in SRCC on LSVQ and \textbf{2-3\%} boost in SRCC on LSVQ-1080p. The gain is larger on LSVQ-1080p because the dataset contains more high-resolution videos, and therefore more quality information is lost when resized statically to a small resolution. Armed with the multi-resolution input representation, \ours\ is able to utilize both global information from lower-resolution frames and detailed information from higher-resolution frames. The results demonstrate that the proposed multi-resolution representation is indeed effective for capturing the complex multi-scale quality information that can be lost when using statically resized frames. Table~\ref{tab:multi-resolution-results} also shows that \ours\ performance improves with the increase of number of input frames since more temporal information is preserved.

After verifying that the multi-resolution representation is indeed more effective than fixed resolution, we also run ablations with different multi-resolution patch sampling methods (Table \ref{tab:patch-sampling-method}). For ``Random", we first resize the frames to the 4-scale multi-resolution input, and then randomly sample the same number of patches from each resolution. For ``High-res Patch on Last Frame", we use low-resolution  patches for the first 3 frames ($224\times$), and only sample high-resolution patches from the last frame ($896\times$).  \ours\ samples center-aligned patches from the 4-scale input, and it performs the best. This shows the proposed sampling method can more effectively utilize the complementary nature of the multi-resolution views. With the center-aligned multi-resolution patches, \ours\ is able to better aggregate both the global view, and the multi-resolution local details.

\begin{table}[!tp]
\centering
\small
\begin{tabular}{ccc|cc}\toprule
&\multicolumn{2}{c|}{LSVQ} &\multicolumn{2}{c}{LSVQ-1080p} \\\cmidrule{2-5}
Patch Sampling Method  &SRCC &PLCC &SRCC &PLCC \\\midrule
Random &0.839 &0.838 &0.739 &0.783 \\
High-res Patch on Last Frame &0.854 &0.855 &0.757 &0.801 \\
\ours &\ourresult{0.867} &\ourresult{0.865} &\ourresult{0.780} &\ourresult{0.817} \\
\bottomrule
\end{tabular}
\caption{Ablation for multi-resolution patch sampling method. Our final method is marked in \noindent\colorbox{gray!30}{\makebox[1em]{gray}}.} \label{tab:patch-sampling-method}
\end{table}

\subsubsection{Number of Grouped Multi-resolution Frames $N$}

\begin{table}[!t]
\centering
\small
\begin{tabular}{cccc|ccc}\toprule
& &\multicolumn{2}{c|}{\ours} &\multicolumn{2}{c}{ w/o Multi-resolution} \\\cmidrule{3-6}
$N$ &GFLOPs &SRCC &PLCC &SRCC &PLCC \\\midrule
2 &534 &0.751 &0.797 &0.742 &0.786 \\
3 &358 &0.757 &0.794 &0.741 &0.786 \\
4 &271 &0.764 &0.802 &0.749 &0.787 \\
5 &218 &0.764 &0.805 &0.743 &0.783 \\
\bottomrule
\end{tabular}
\caption{Ablation study results for number of grouped frames $N$ on the LSVQ-1080p dataset. \ours\ uses multi-resolution input while ``w/o Multi-resolution" use fixed resolution frames. Models here are trained with 60 input frames instead of 128. } \label{tab:number-of-scales-results}
\end{table}

\noindent In Table~\ref{tab:number-of-scales-results} we run ablations on the number of grouped frames $N$ when building the multi-resolution video representation. The experiment is run with 60 frames instead of 128 since smaller $N$ increases the number of input tokens for the temporal encoder and introduces high computation and memory cost. For \ours, we use multi-resolution input for the grouped frames and for ``w/o Multi-resolution", we resize all the frames to the same 224 shorter-side length. For all $N$, using multi-resolution input is better than a fixed resolution. It further verifies the effectiveness of the proposed multi-resolution input structure. For multi-resolution input, the performance improves when increasing $N$ from 2 to 5, but the gain becomes smaller as $N$ grows larger. There is also a trade-off between getting higher resolution views and the loss of spatio-temporal information with the increase of $N$, since the area ratio of sampled patches becomes smaller as resolution increases Overall, we find $N=4$ to be a good balance between performance and complexity.

\begin{table}[!t]
\centering
\small
\begin{tabular}{ccc|cc|ccc}\toprule
& & &\multicolumn{2}{c|}{LSVQ} &\multicolumn{2}{c}{LSVQ-1080p} \\\cmidrule{4-7}
Pretrain Dataset &\#Images &Task &SRCC &PLCC &SRCC &PLCC \\\midrule
ImageNet~\citep{imagenet} &1M &Class. &0.839 &0.837 &0.748 &0.780 \\
ImageNet-21k~\citep{deng2009imagenet} &14M &Class. &\ourresult{0.867} &\ourresult{0.865} &\ourresult{0.780} &\ourresult{0.817} \\
LIVE-FB~\citep{ying2020patches} &160K &IQA &0.848 &0.846 &0.760 &0.788 \\
\bottomrule
\end{tabular}
\caption{Results for initializing \ours\ model from checkpoints pretrained on different image datasests. Our final method is marked in \noindent\colorbox{gray!30}{\makebox[1em]{gray}}.} \label{tab:pretrained-dataset-results}
\end{table}

\begin{table}[!t]
\centering
\small
\begin{tabular}{crr|rrr}\toprule
&\multicolumn{2}{c|}{LSVQ} &\multicolumn{2}{c}{LSVQ-1080p} \\\cmidrule{2-5}
Frame Sampling Method &SRCC &PLCC &SRCC &PLCC \\\midrule
Uniform Sample &\ourresult{0.867} &\ourresult{0.865} &\ourresult{0.780} &\ourresult{0.817} \\
Front Sample &0.860 &0.857 &0.773 &0.808 \\
Center Clip &0.860 &0.857 &0.771 &0.811 \\
\bottomrule
\end{tabular}
\caption{Ablation study results for frame sampling method. Our final method is marked in \noindent\colorbox{gray!30}{\makebox[1em]{gray}}.} \label{tab:frame-sampling-method}
\end{table}

\subsubsection{Pretrained Checkpoint Selection} 
\label{sec:prerained-dataset-results}
\noindent Compared to CNNs, Transformers impose less restrictive inductive biases which broadens their representation ability. On the other hand, since Transformers lack the inductive biases of the 2D image structure, it generally needs large datasets for pretraining to learn the inductive priors. In Table~\ref{tab:pretrained-dataset-results}, we try initializing the spatial Transformer encoder in \ours\ model with checkpoints pretrained on different image datasets, including two image classification (Class.) datasets, and one image quality assessment (IQA) dataset. ImageNet-21k is the largest and it performs the best, showing that large-scale pretraining is indeed beneficial. This conforms with the findings in previous vision Transformer works \citep{arnab2021vivit, dosovitskiy2021an}. LIVE-FB~\citep{ying2020patches} is an IQA dataset on which PVQ~\citep{ying2021patch} obtain their 2D frozen features. Since IQA is a very relevant task to VQA, pretraining on this relatively small IQA dataset leads to superior results than ImageNet. This shows that relevant task pretraining is beneficial when large-scale pretraining is not accessible.

\subsubsection{Frame Sampling Strategy} 
\noindent We run ablations on the frame sampling strategy in Table~\ref{tab:frame-sampling-method}. For our default ``Uniform Sample", we sample 128 frames uniformly throughout the video. For ``Front Sample", we sample the first 128 frames. For ``Center Clip" we take the center clip of 128 frames from the video. On LSVQ and LSVQ-1080p dataset, uniformly sampling the frames is the best probably because there is temporal redundancy between continuous frames and uniformly sampling the frames allows the model to see more diverse video clips. Since most of the videos in the VQA dataset are relatively short, uniformly sampling the frames is good enough to provide a comprehensive view. 

\section{Conclusion and Future Work}

\noindent We propose a multi-resolution Transformer (\ours) for VQA, which integrates multi-resolution views to capture both global and local quality information. By transforming the input frames to a multi-resolution representation with both low and high resolution frames, the model is able to capture video quality information at different granularities. To effectively handle the variety of resolutions in the multi-resolution input sequence, we propose a multi-resolution patch sampling mechanism. A factorization of spatial and temporal Transformers is employed to efficiently model spatial and temporal information and capture complex space-time distortions in UGC videos. Experiments on several large-scale UGC VQA datasets show that \ours\ can achieve state-of-the-art performance and has strong generalization capability, demonstrating the effectiveness of the proposed method. \ours\ is designed for VQA, and it can be extended to other scenarios where the task labels are affected by both video global composition and local details. The limitation of Transformers is that it can be computationally expensive, and thus costly to make predictions on long videos. In this paper, we focus on improving the performance of the VQA model and we leave it as future work to improve its efficiency and to lower the computation cost. One potential direction is to employ more efficient Transformer variants, such as Reformer~\citep{kitaev2020reformer} and Longformer~\citep{beltagy2020longformer} where the attention complexity has been greatly reduced. Those efficient Transformers can be adopted as a drop-in replacement for the current spatial and the temporal Transformer used in \ours. 

\bibliographystyle{Frontiers-Harvard} 
\bibliography{main}


\end{document}